\documentclass[letterpaper]{article} 
\usepackage{aaai23}
\usepackage{float}
\usepackage{times}  
\usepackage{helvet}  
\usepackage{courier}  
\usepackage[hyphens]{url}  
\usepackage{graphicx} 
\usepackage{natbib}  
\usepackage{caption} 
\usepackage{amsmath,amssymb,stmaryrd}  
\usepackage{algorithm}
\usepackage{algpseudocode}
\usepackage{multirow}
\usepackage[nodisplayskipstretch]{setspace}
\usepackage{array, multirow}
\usepackage{newfloat}
\usepackage{listings}
\DeclareCaptionStyle{ruled}{labelfont=normalfont,labelsep=colon,strut=off} 
\floatstyle{ruled}
\newfloat{listing}{tb}{lst}{}
\floatname{listing}{Listing}
\title{Latents2Semantics: Leveraging the Latent Space of Generative Models for Localized Style Manipulation of Face Images}
\author {
    Snehal Singh Tomar, A.N. Rajagopalan 
}
\affiliations {
    Indian Institute of Technology Madras\\
    snehal@smail.iitm.ac.in, raju@ee.iitm.ac.in
}

\usepackage{bibentry}

\begin{document}

\maketitle

\begin{abstract}
With the metaverse slowly becoming a reality and given the rapid pace of developments toward the creation of digital humans, the need for a principled style editing pipeline for human faces is bound to increase manifold. We cater to this need by introducing the Latents2Semantics Autoencoder (L2SAE), a Generative Autoencoder model that facilitates highly localized editing of style attributes of several Regions of Interest (ROIs) in face images. The L2SAE learns separate latent representations for encoded images' structure and style information. Thus, allowing for structure-preserving style editing of the chosen ROIs. The encoded structure representation is a multichannel 2D tensor with reduced spatial dimensions, which captures both local and global structure properties. The style representation is a 1D tensor that captures global style attributes. In our framework, we slice the structure representation to build strong and disentangled correspondences with different ROIs. Consequentially, style editing of the chosen ROIs amounts to a simple combination of (a) the ROI-mask generated from the sliced structure representation and (b) the decoded image with global style changes, generated from the manipulated (using Gaussian noise) global style and unchanged structure tensor. Style editing sans additional human supervision is a significant win over SOTA style editing pipelines because most existing works require additional human effort (supervision) post-training for attributing semantic meaning to style edits. We also do away with iterative-optimization-based inversion or determining controllable latent directions post-training, which requires additional computationally expensive operations. We provide qualitative and quantitative results for the same over multiple applications, such as selective style editing and swapping using test images sampled from several datasets.   
\end{abstract}

\section{Introduction}
\label{sec:intro}
With the rise of online photo-sharing applications, the demand for tools that enable automatic photorealistic edits on face images has seen exponential growth. Motivated by the capability of Generative Adversarial Networks (GANs) to generate high-resolution and photorealistic images and transfer the style attributes from an image to another (demonstrated by works like \cite{DBLP:journals/corr/abs-1710-10196, DBLP:journals/corr/abs-1812-04948, Karras_2020_CVPR}), 
\begin{center}
\begin{figure}[t]
  \centering
        \includegraphics[width = 0.45\textwidth]{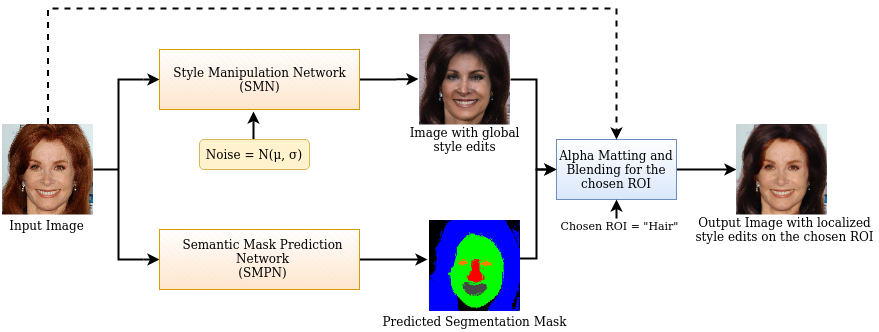} 
        \caption{A high level representation of our ROI-specific style editing pipeline and its constituents.}
        \label{fig:overview}
\end{figure}
\end{center}
several GAN-based approaches \cite{Zhu_2020_CVPR, alaluf2021matter, Nizan_2020_CVPR, wang2021cross, d'Apolito_2021_CVPR, pmlr-v119-nie20a, shen2020interfacegan, Patashnik_2021_ICCV, NEURIPS2020_6fe43269, Niemeyer2020GIRAFFE} have attempted this task. However, there exists a paucity of methods that can efficiently perform highly localized, structure-preserving, and photorealistic style edits on real images which are in accordance with the global style scheme. Very recently, a few methods like \cite{Shi_2022_CVPR} addressed local style editing of real images, but their framework is dependent on the highly expensive iterative optimization-based image-to-latent inversion process \cite{alaluf2021restyle} and thus inconvenient for real-life applications.

\par We seek to diminish the aforesaid research gap in this work. The latent space of Generative Autoencoder (GA) models can be designed to encode the \textit{structure} and \textit{style} information present in the input image into separate representations. Our key insight is that achieving a strong correspondence between a part of the structure representation and semantic ROIs in a disentangled fashion is pivotal for solving the problem. To this end, we design a framework where we slice structure representation such that each slice represents (gets interpreted as) all the information required by the decoder to decode each ROI independently. The style representation can then be leveraged to affect global style changes onto the input image, and the sliced structure representation can be leveraged to produce semantic segmentation masks for each ROI. Consequentially, obtaining ROI-specific style edits shall amount to a simple alpha matting task with the predicted semantic segmentation masks as the matte.

Furthermore, our design is based on Swapping Autoencoder (SAE) \cite{park2020swapping} that learns a latent representation with a neat distinction between structure and style information of the input image. In contrast to iterative-optimization-based inversion approaches, our design is significantly faster, as it inverts using a single forward pass and converges to areas of the latent space, which are more suitable for editing. We demonstrate that our simple yet effective autoencoder-based approach achieves satisfactory visual quality while being extremely fast compared to prior art \cite{Shi_2022_CVPR} for local style editing of face images. The contributions of this paper can be summarized as below:
\begin{itemize}
    \item We employ the SAE's latent space consisting of the Structure Tensor and Texture (Style) Vector to perform localized, structure-preserving, and photorealistic style edits on real face images, which are all independent of one another and are in accordance with the global style scheme of the input image. Our approach does not require any priors or ground-truth (GT) supervision for the locally altered images, peripheral models, or additional data for editing an input image. To our knowledge, ours is the first work that does so for any Generative Autoencoder model; eliminating the need for expensive inversions. We provide qualitative results and quantitative metrics to support our claims.   
    \item We infuse a strong disentanglement with respect to the structure of semantic ROIs in re-generated images in the latent space (Structure Tensor) of SAE \cite{park2020swapping}. Our model achieves satisfactory visual quality compared to SOTA while being extremely fast and efficient.  
     \item Localized style editing for a given ROI from an input image amounts to a simple forward pass with the addition of noise to the texture (style) latents for achieving global style edits, followed by a forward pass with appropriate masking (retention of a single non-zero slice corresponding to the ROI) applied to the structure latents for predicting ROI specific segmentation masks. Finally, an alpha matting operation generates the output image. This eliminates the need for any additional human effort (supervision) post-training and underscores the applicability of our model to any generic automatic face editing application.  
\end{itemize}

\section{Method}
An overview of our framework is shown in Fig. \ref{fig:overview}. Our ROI-specific style editing pipeline consists of two primary networks, viz., the Style Manipulation Network (SMN) and the Semantic Mask Prediction Network (SMPN). Given an image, the SMN is tasked with generating photorealistic global style modified images, and the SMPN is responsible for generating accurate segmentation maps. The Alpha matting block performs an alpha matting between the input image and the style modified image using the predicted ROI-specific segmentation mask as the matte. This results in the output image's style attributes being modified only for the chosen ROI. A key advantage of our method is that it does not require GT supervision for the locally-modified images and is prior agnostic. During training, it only needs 
\begin{center}
\begin{figure}[ht]
  \centering
        \includegraphics[width = 0.47\textwidth]{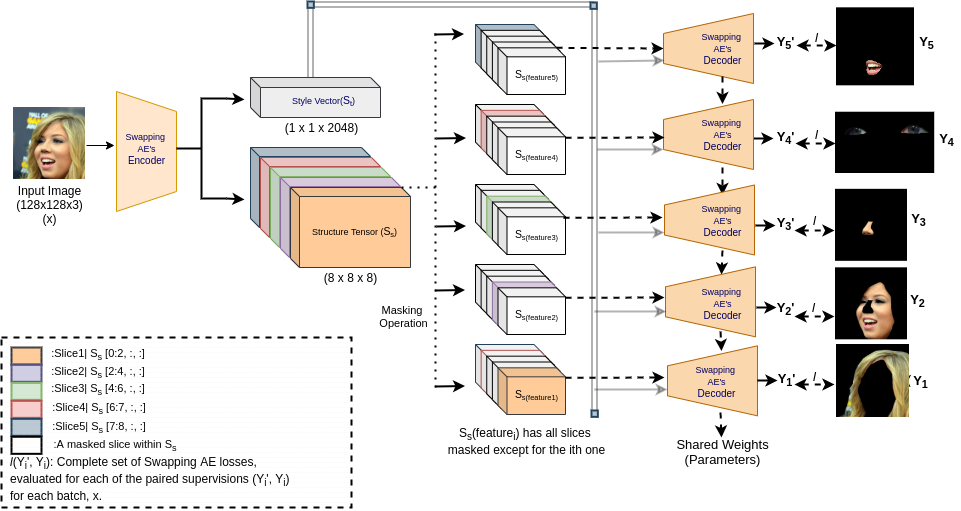} 
        \caption{A schematic representation of the SAE \cite{park2020swapping} inspired SMN and SMPN architectures and training methodology. The schema used for slicing $S_{s}$ and for formation of masked $S_{s_{feature_{i}}}$ have been annotated in the legend. $l$ refers to the operation defined by Eq. 1.}
        \label{fig:scheme}
\end{figure}
\end{center}
the weak supervision of the GT segmentation maps as guidance. The SMN and the SMPN share the same model architecture, derived from SAE \cite{park2020swapping}. The SAE \cite{park2020swapping} is a generative Autoencoder model which embeds the \textit{structure} and \textit{style} information present in input images ($H \times H \times 3$) into a \textit{structure tensor} ($S_{s}$, having dimensions $H/16 \times H/16  \times 8$) and \textit{texture vector} ($S_{t}$, having dimensions $1 \times 1 \times 2048$). The latent space $S = \{S_{s}, S_{t}\}$ serves as the foundation of our work. 
\newline Our primary goal is to manipulate the style of a particular region in the image. To achieve that, we first aim to build a correspondence between the latent representation $S_{s}$ and the structure of individual regions of interest, namely: hair, skin, nose, eyes, and (lips + mouth) in the reconstructed image. In our work, we follow a feature slicing scheme in the latent space and found it to perform reasonably well in practice. For each ROI, we first fix a particular set of feature maps of $S_{s}$ and mask (setting to zero) all other feature slices. Next, we train the network to produce an image containing only the corresponding semantic ROI when the masked $S_{s}$ is decoded together with $S_{t}$. 

\begin{algorithm}
\hspace*{\algorithmicindent} \textbf{Input}: $x$ (input image), $Choice_{ROI}, \mu_{stlye-noise}$ \\
\hspace*{\algorithmicindent} \textbf{Output}: $\hat{x}$ (ROI-selective style edited image)
\caption{Inference algorithm}\label{alg:euclid}
\begin{algorithmic}[1]
\State $SMN = SAE(), SMPN \gets SAE_{sliced_{S_{s}}}()$
\State $noise \gets \mu_{stlye-noise} \cdot \mathbb{N}(0, 1)$, \\$slice\_mask \gets Mask_{Choice_{ROI}}$ 
\State $S_{s_{1}}, S_{t_{1}} = Encoder_{SMN}(x)$
\State $NoisyS_{t_{1}} = S_{t_{1}} + noise$
\State $x_{Globally Noisy} = Decoder_{SMN}(S_{s_{1}}, NoisyS_{t_{1}})$ 
\State $S_{s_{2}}, S_{t_{2}} = Encoder_{SMPN}(x)$
\State $Sroi_{s_{2}} = slice\_mask \cdot S_{s_{2}}$
\State $ROI_{Mask} = Decoder_{SMPN}(Sroi_{s_{2}}, S_{t_{2}}) > 0$ 
\State $\hat{x} = AlphaMatting(x, x_{Globally Noisy}, ROI_{SemanticMask})$ \Comment{Operation defined by Algorithm \ref{algo: matting}}
\end{algorithmic}
\label{algo: infer}
\end{algorithm}
\begin{algorithm}[t]
\hspace*{\algorithmicindent} \textbf{Input}: $x$ (input image), $m$ (ROI mask), $y$ (global style edited image)\\
\hspace*{\algorithmicindent} \textbf{Output}: $\hat{x}$ (ROI-selective style edited image)
\caption{The alpha matting and blending algorithm}\label{alg:euclid}
\begin{algorithmic}[1]
\State $\alpha = m \circledast \sigma(3,3)$ \Comment{$\circledast$ denotes convolution,  $\sigma(i, j) = (1/\sqrt{2\pi}) \cdot e^{-(i^{2} + j^{2})/2}; i, j \in [0, 3)$}
\State $\hat{x} = (1 - \alpha) \cdot x + \alpha \cdot y$
\State $\hat{x} = \hat{x} \circledast \sigma(3,3)$ \Comment{$\circledast$ denotes convolution}
\end{algorithmic}
\label{algo: matting}
\end{algorithm}
\begin{figure*}[t]
    \centering
    \resizebox{13cm}{!}{
      \begin{tabular}{c c c c c c c}
        \tiny Input & \tiny SMPN Output & \tiny Hair & \tiny Eyes & \tiny Nose & \tiny Lips + Mouth & \tiny Skin \\
        \includegraphics[scale = 0.0625]{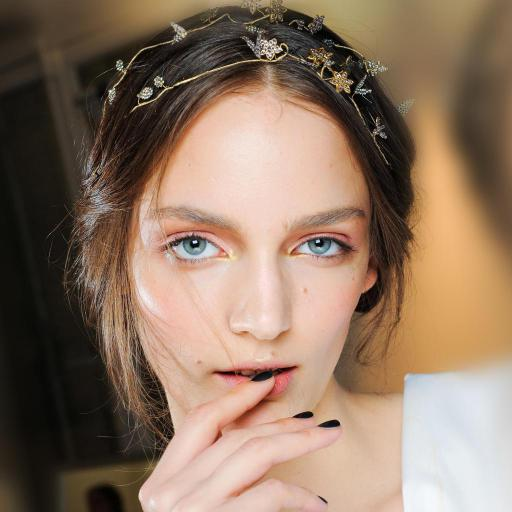} & \includegraphics[scale = 0.25]{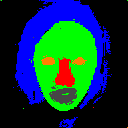} & \includegraphics[scale = 0.25]{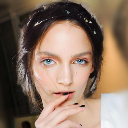} & \includegraphics[scale = 0.25]{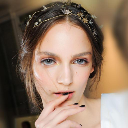} & \includegraphics[scale = 0.25]{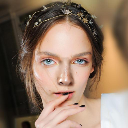} & \includegraphics[scale = 0.25]{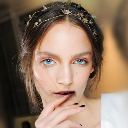} & \includegraphics[scale = 0.25]{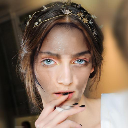}\\
        \includegraphics[scale = 0.0625]{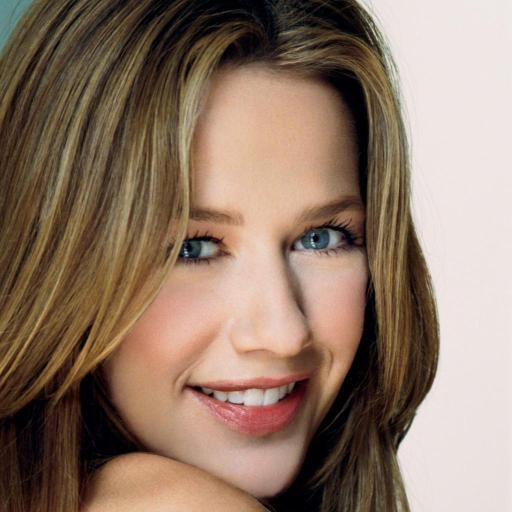} & \includegraphics[scale = 0.25]{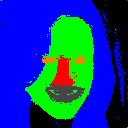} & \includegraphics[scale = 0.25]{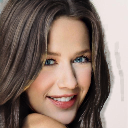} & \includegraphics[scale = 0.25]{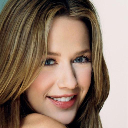} & \includegraphics[scale = 0.25]{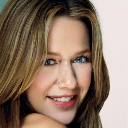} & \includegraphics[scale = 0.25]{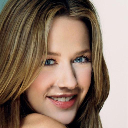} & \includegraphics[scale = 0.25]{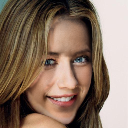}     
    \end{tabular}}
    \caption{Qualitative results for the prediction of semantic segmentation maps (column 2) by the SMPN and selective style edits (columns 3 through 7) performed on several ROIs. All selective style editing results were obtained by giving noise vectors sampled from $\mathbb{N}$(0, 1) as input to the SMN. The color coding used for semantic regions in the segmentation maps is given by; blue: \textit{hair}, green: \textit{skin}, red: \textit{nose}, orange (approach 1): \textit{eyes}, orange (approach 2): \textit{lips + mouth}, grey (approach 1): \textit{lips + mouth}, and grey (approach 2): \textit{eyes} .}
    \label{fig: just_qual}
\end{figure*}

\begin{figure}[h]
    \centering
    \resizebox{8.5cm}{!}{
      \begin{tabular}{c c c c c c}
        & \tiny Eyes & \tiny Hair & \tiny Nose & \tiny Lips + Mouth & \tiny Skin\\
         \tiny{Style Reference}& \includegraphics[scale = 0.0625]{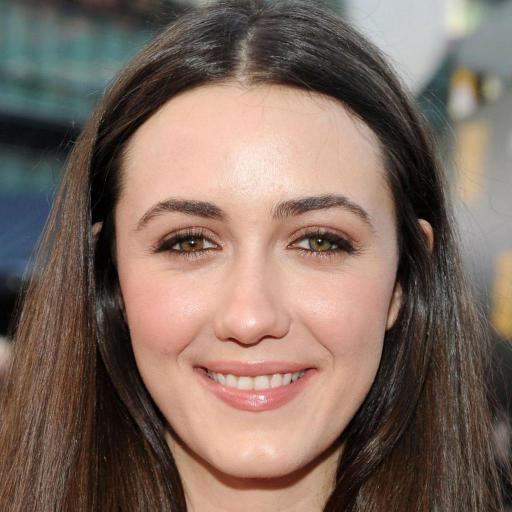} & \includegraphics[scale = 0.0625]{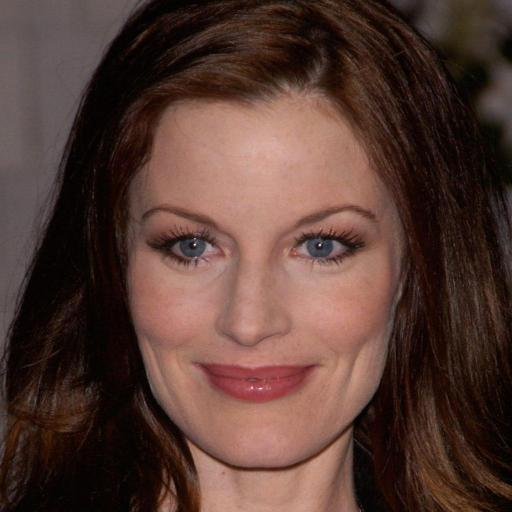} & \includegraphics[scale = 0.0625]{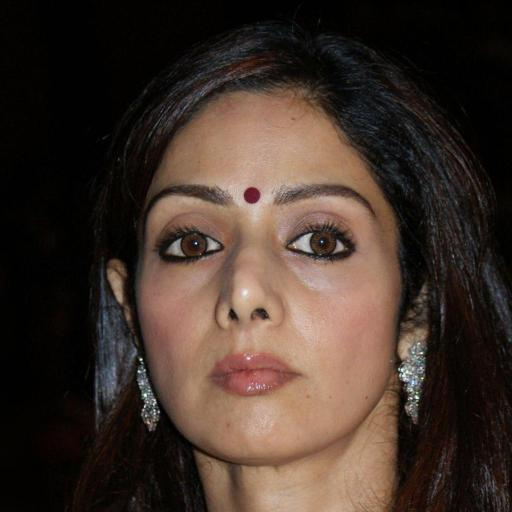} & \includegraphics[scale = 0.0625]{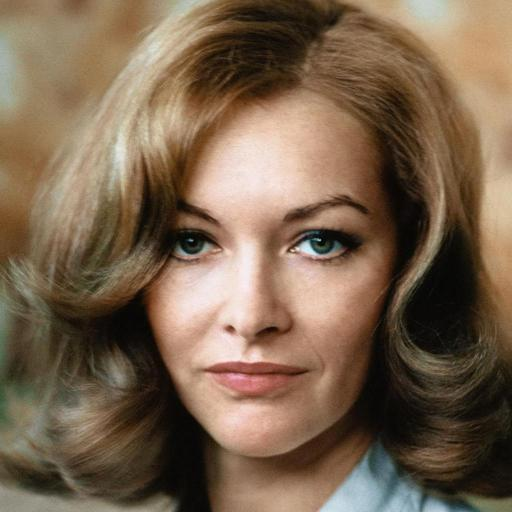} & \includegraphics[scale = 0.0625]{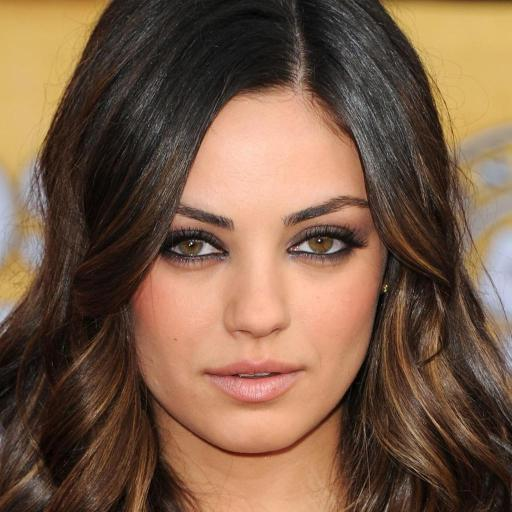} \\
        \includegraphics[scale = 0.0625]{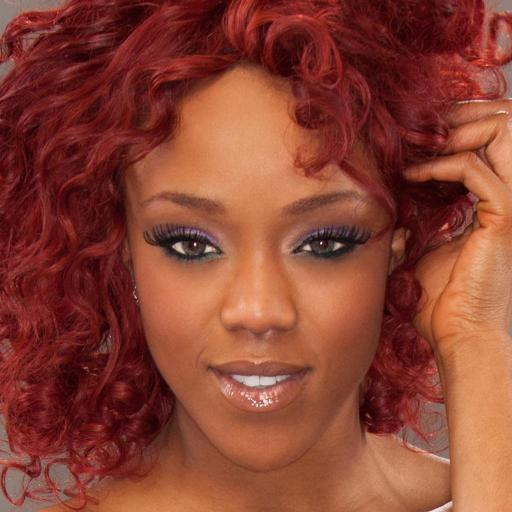} & \includegraphics[scale = 0.25]{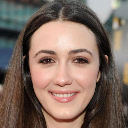} & \includegraphics[scale = 0.25]{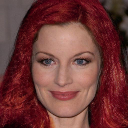} & \includegraphics[scale = 0.25]{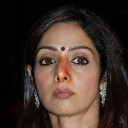} & \includegraphics[scale = 0.25]{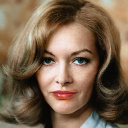} & \includegraphics[scale = 0.25]{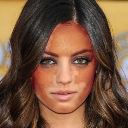}\\
        \includegraphics[scale = 0.0625]{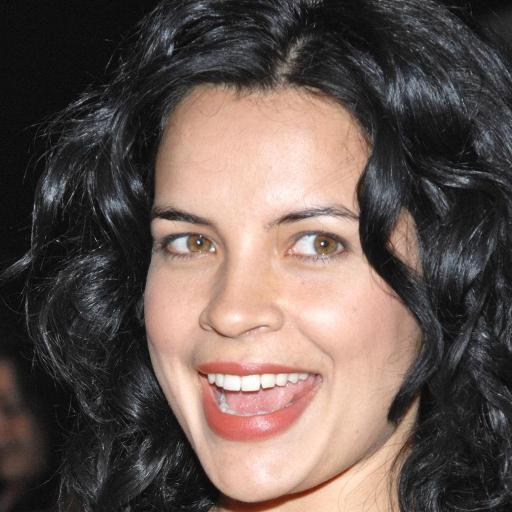} & \includegraphics[scale = 0.25]{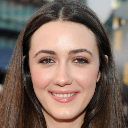} & \includegraphics[scale = 0.25]{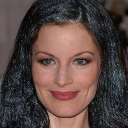} & \includegraphics[scale = 0.25]{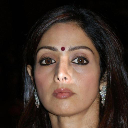} & \includegraphics[scale = 0.25]{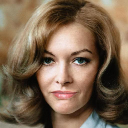} & \includegraphics[scale = 0.25]{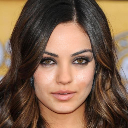}\\
\end{tabular}}
\caption{Selective style swapping across several ROIs. Column 1 depicts images used as a style reference. Columns 2 through 6 in the top row show input images (structre reference) for performing style edits. Images in the matrix denote the outputs for selective swapping (with respect to the ROI denoted by the column) of style attributes between the corresponding input and style reference image, respectively.}
\label{fig: sel_style_swapping}
\end{figure}

\begin{figure}[h]
    \centering
    \resizebox{6.5cm}{!}{
      \begin{tabular}{c c c p{1cm}}
        \tiny ROI & \tiny Input & \tiny Reconstruction & \tiny Manipulated Reconstruction\\
         \tiny{Hair}& \includegraphics[scale = 0.0625]{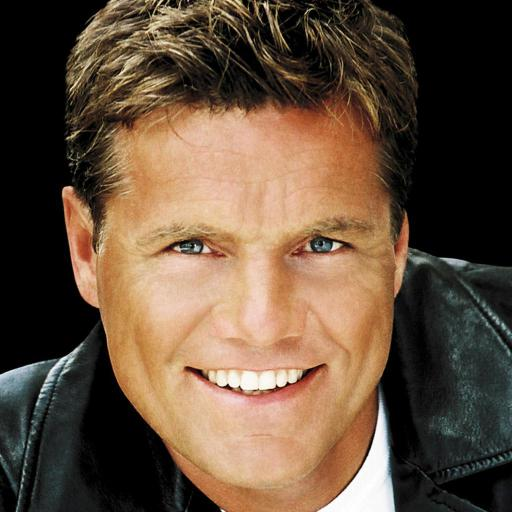} & \includegraphics[scale = 0.25]{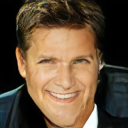} & \includegraphics[scale = 0.25]{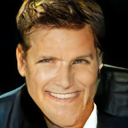} \\ 
         \tiny{Skin}& \includegraphics[scale = 0.0625]{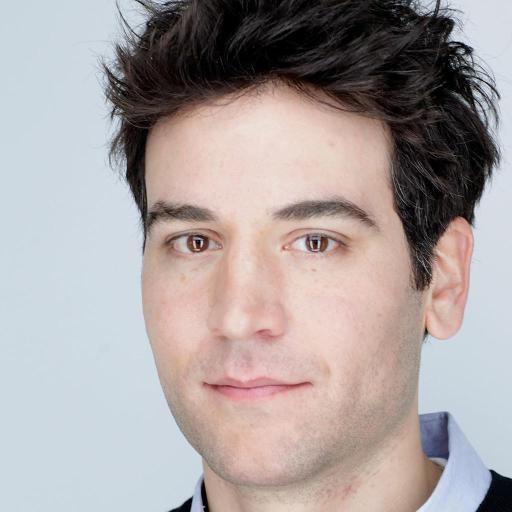} & \includegraphics[scale = 0.25]{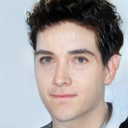} & \includegraphics[scale = 0.25]{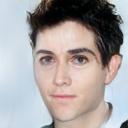} \\
         \tiny{Lips}& \includegraphics[scale = 0.0625]{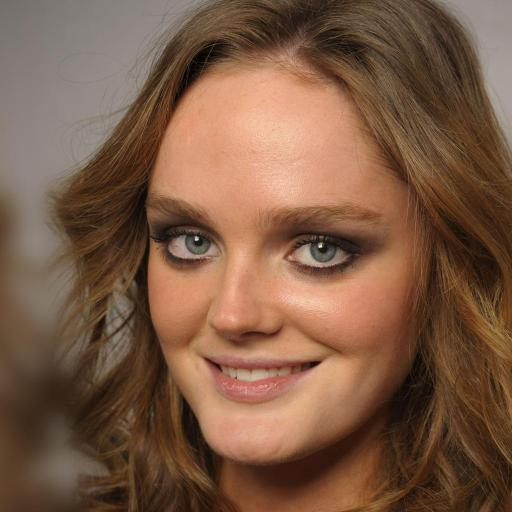} & \includegraphics[scale = 0.25]{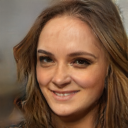} & \includegraphics[scale = 0.25]{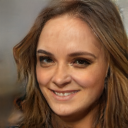} \\
\end{tabular}}
\caption{ROI-wise structure editing. Column 2 depicts the reconstructed image obtained sans any modification to encoded latent. Column 3 shows the structure modifications obtained upon reconstruction with noise added to ROI-specific-slices of the encoded latents.}
\label{fig: struct_edit}
\end{figure}
We use separate network instances for the SMN and  the SMPN. The SMN is a pre-trained SAE (trained on the FFHQ dataset \cite{DBLP:journals/corr/abs-1812-04948}). Whereas, the SMPN is a separate SAE instance, fine-tuned on the ROI-separated CelebAMask-HQ dataset comprising 27016 datapoints. 
\paragraph{Network Architecture and Losses.} 
\begin{equation}
\label{eq:loss_for_global}
\begin{split}
      l(Y_{0}', Y_{0}) = L_{\text{rec}}(Y_{0}', Y_{0}) & + 0.5L_{\text{GAN, rec}}(Y_{0}', Y_{i})\\ +  0.5L_{\text{GAN, swap}}(Y_{0}', Y_{0}) &  + 0.5L_{\text{CooccurGAN}}(Y_{0}', Y_{0})
\end{split}
\end{equation}
The pretrained SAE \cite{park2020swapping} is taken such that it id trained on the FFHQ dataset  using the same loss functions as Eq. \ref{eq:loss_for_global}. Given an image, it first generates $S_s$ and $S_t$. Any perturbation in the style tensor $S_t$ results in global style manipulation in the reconstructed image.  
\par The SMPN solely focuses on segmenting different ROIs given the input image. We follow the same architecture and slicing scheme of the structure latent $S_s$ as elucidated in Fig. \ref{fig:scheme}. For every batch of training data, parameters of the encoder and decoder (all Siamese decoders share the same parameters) are optimized using the following overall loss:
\begin{equation}
\label{eq:overall_app2}
    L_{\text{overall}} = \sum_{i = 1}^{5} 0.2 \cdot l(Y_{i}', Y_{i})
\end{equation}
Overall, the SMN produces an image with a different global style. The ROI masks, predicted by the SMPN, are used to allow style changes only in certain semantic regions using alpha blending.
\subsubsection{Alpha Matting and Blending}
Given the input image ($x$), the semantic mask for the chosen ROI ($m$), and the global style edited image ($y$). The $AlphaMatting()$ operation used in Algorithm \ref{algo: infer} to obtain the ROI localized style edited image ($\hat{x}$) corresponds to the series of steps given by Algorithm \ref{algo: matting}, where $\sigma(3,3)$ denotes a standard $3 \times 3$ Gaussian convolution kernel. The Gaussian blurring performed, ensures smoothing of edges in the matte and combined image, respectively.
\subsection{Training}
Training was initiated using the pre-trained weights provided by    \cite{park2020swapping}, post training on the FFHQ dataset \cite{DBLP:journals/corr/abs-1812-04948}. The optimizer and training loop used were the same as used by \cite{park2020swapping}. As show in Figure 3, a batch of training data comprised $\{X, Y_{1}, ..., Y_{5}\}$ where $X$ denotes a batch of input images and $Y_{i}$ denotes a batch of region specific images, $R_{i}$. 
\begin{table}
\centering
\caption{Quantitative results for our methods performance versus SemanticStyleGAN \cite{Shi_2022_CVPR}. We evaluate the perceptual similarity of edits obtained (FID, LPIPS) and the time taken to perform one texture edit per ROI.}
\resizebox{8cm}{!}{
\begin{tabular}{|c|c c c|}
 \hline
 \textbf{Method} & FID $\downarrow$ & LPIPS $\downarrow$ & Time Taken (s) $\downarrow$\\
  \hline
  \textbf{SemanticStyleGAN}  & 0.3072 & \textbf{22.3771} & 120.602\\
  \textbf{Approach 2 (Ours)} & \textbf{0.2026} & 26.6084 & \textbf{0.07}\\
 \hline 
\end{tabular}}
\label{tab: quant}
\end{table}
The SMPN model was trained on the CelebAMask-HQ dataset \cite{CelebAMask-HQ}.  
\subsection{Inference Mechanism}
Inferring segmentation maps from the model amounts to a simple forward pass over the network with appropriate masking applied to $S_{s}$ in the latent space, depending upon the ROI for which the segmentation has to be performed. Similarly, performing global style edits entails adding noise to the encoded $S_{t}$ followed by decoding it in tandem with unchanged $S_s$. The process of performing ROI specific style edits using our model is encapsulated by Algorithm \ref{algo: infer}.

\section{Experiments}
\label{sec:exp}
We present a detailed analysis of our model's efficacy towards the claims made in this paper. We provide qualitative and quantitative results for our method's performance with respect to degree of photorealism, ROI-wise localization of edits, and time of computation per edit. The SemanticStyleGAN \cite{Shi_2022_CVPR} is one of the few recent works that claim to perform highly localized texture editing of real images with minimal human supervision post training. We present a detailed comparative study between our method and the SemanticStyleGAN \cite{Shi_2022_CVPR}. The subsequent subsections elucidate that our method is much faster while being comparably good at performing ROI-wise edits. We do not attempt to estimate the amount of additional human effort (supervision) required by competing methods for the sake of brevity. Figures 4 and 5 illustrate selective style swapping and structure editing which are promising applications of our method.

\subsection{Dataset and Implementation Details}
Our model was trained only on the CelebAMask-HQdataset \cite{Karras2018} and was evaluated on 862 images sampled from it. The models were trained on 2 NVIDA GeForce RTX 3090 GPUs using a batch size of 4, respectively. We used the optimizer presented by the SAE \cite{park2020swapping} to train our models. 
\subsection{L2SAE versus SOTA}
\subsubsection{Computation Time for obtaining edits}
Owing to the optimization-based pipelines for inverting real images to latents, most SOTA methods such as \cite{Shi_2022_CVPR} and \cite{Wu_2021_CVPR} require heavy computations for projecting images onto their latent space. These methods lack neat disentanglement with respect to semantic ROIs in their latent manifold. Thus, making editing via inverted latents time consuming. Most of this time is lost in additional human supervision required for attributing meaning to controllable directions and inferring from latent space classification models. We ignore the effects of additional human supervision in our study. Given its disentangled  latent space, our method is much faster than the SOTA in producing ROI-specific edits. Table \ref{tab: quant} shows that our approach is faster than SOTA by multiple orders of magnitude.    
\subsubsection{Quality of Style Edits}
\begin{figure}[h]
    \centering
    \resizebox{13cm}{!}{
      \begin{tabular}{c c c c c c c}
         Method & Input & Hair & Nose &  Skin & Lips + Mouth  & Eyes\\
         \begin{tabular}{c}
              SemanticStyleGAN \\
               \\
              \\ 
         \end{tabular} & \includegraphics[scale = 0.125]{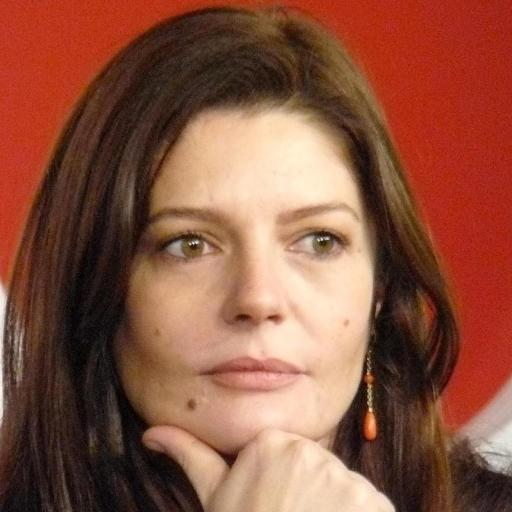} & \includegraphics[scale = 0.07]{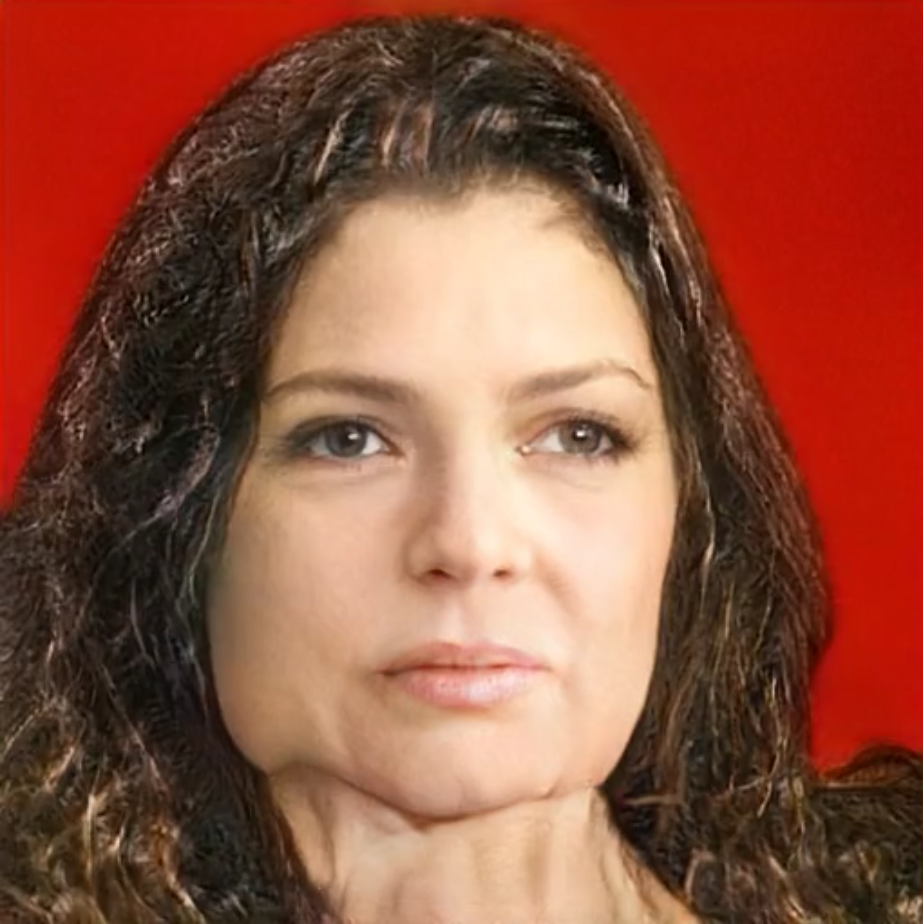} & \includegraphics[scale = 0.07]{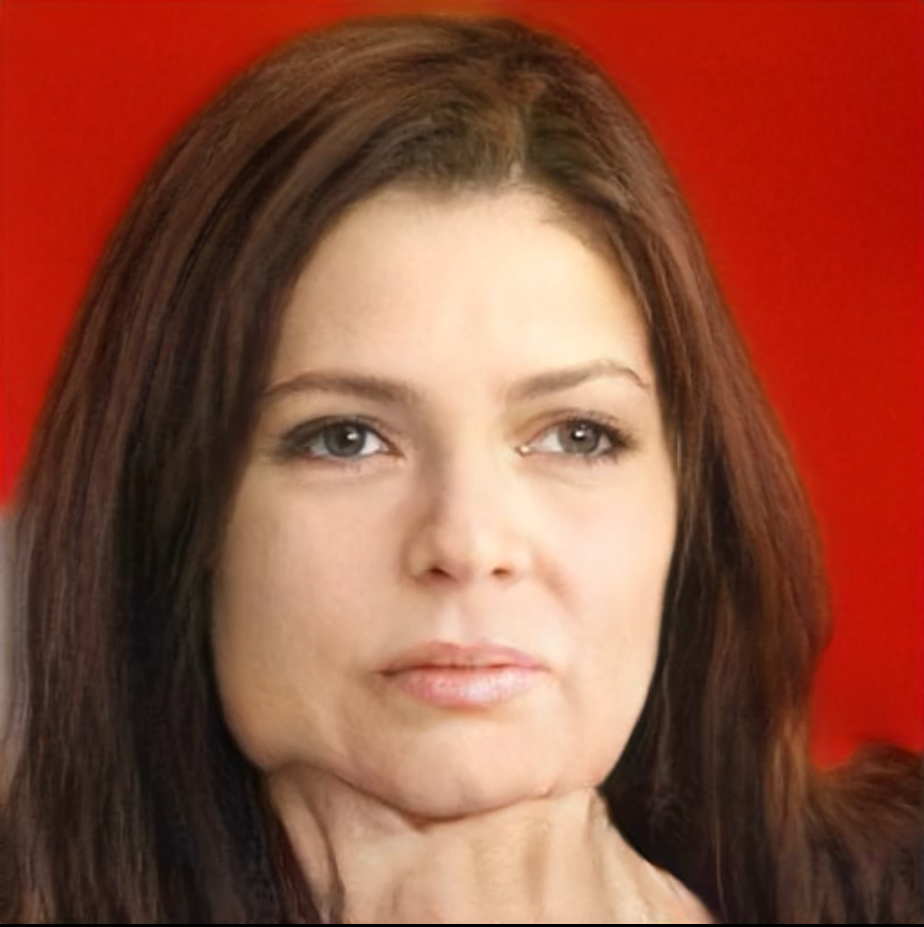} & \includegraphics[scale = 0.07]{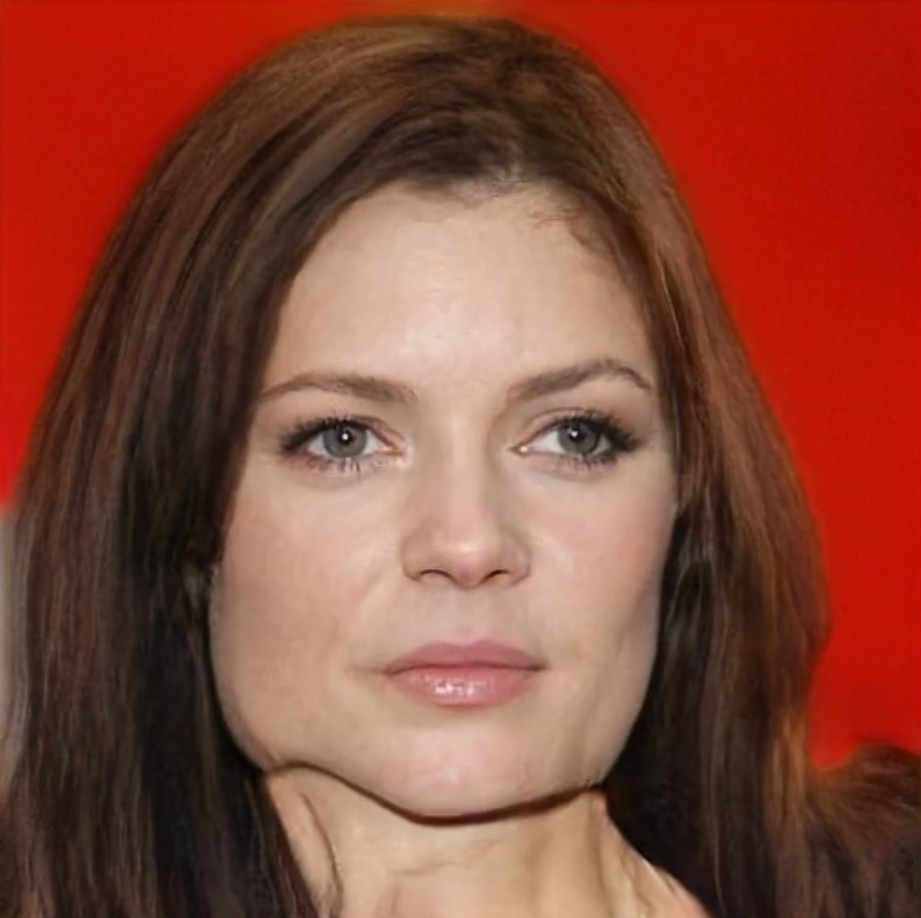} & \includegraphics[scale = 0.07]{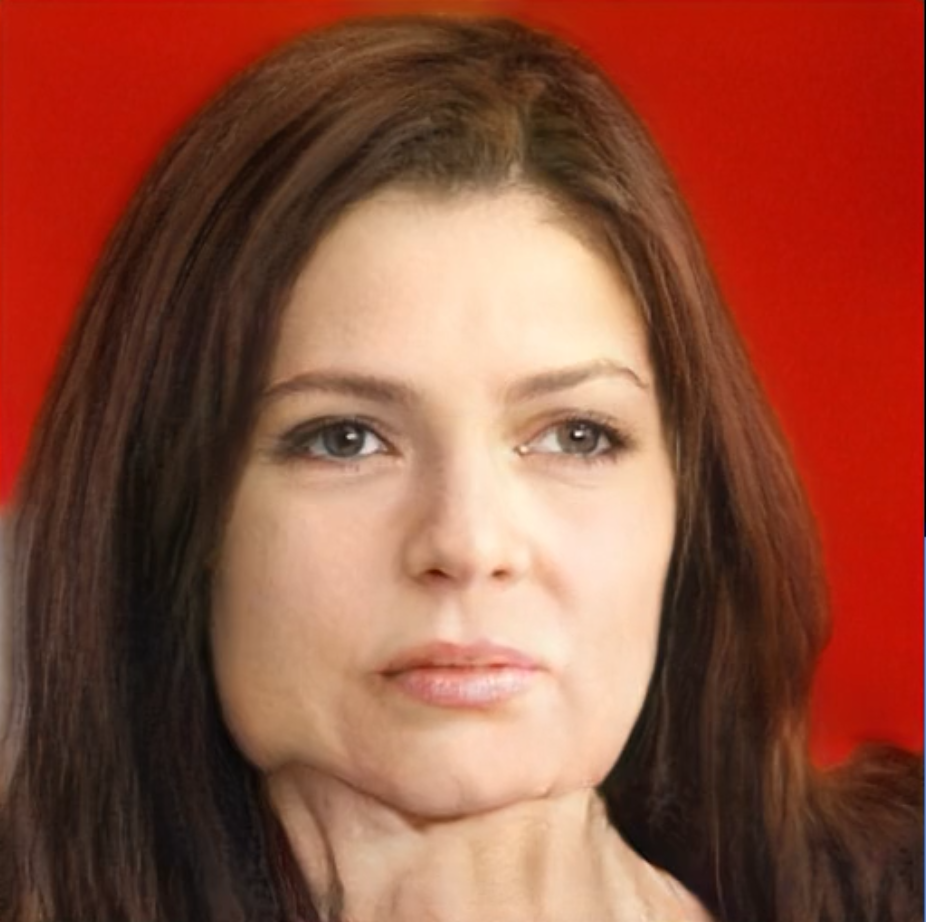} & \includegraphics[scale = 0.07]{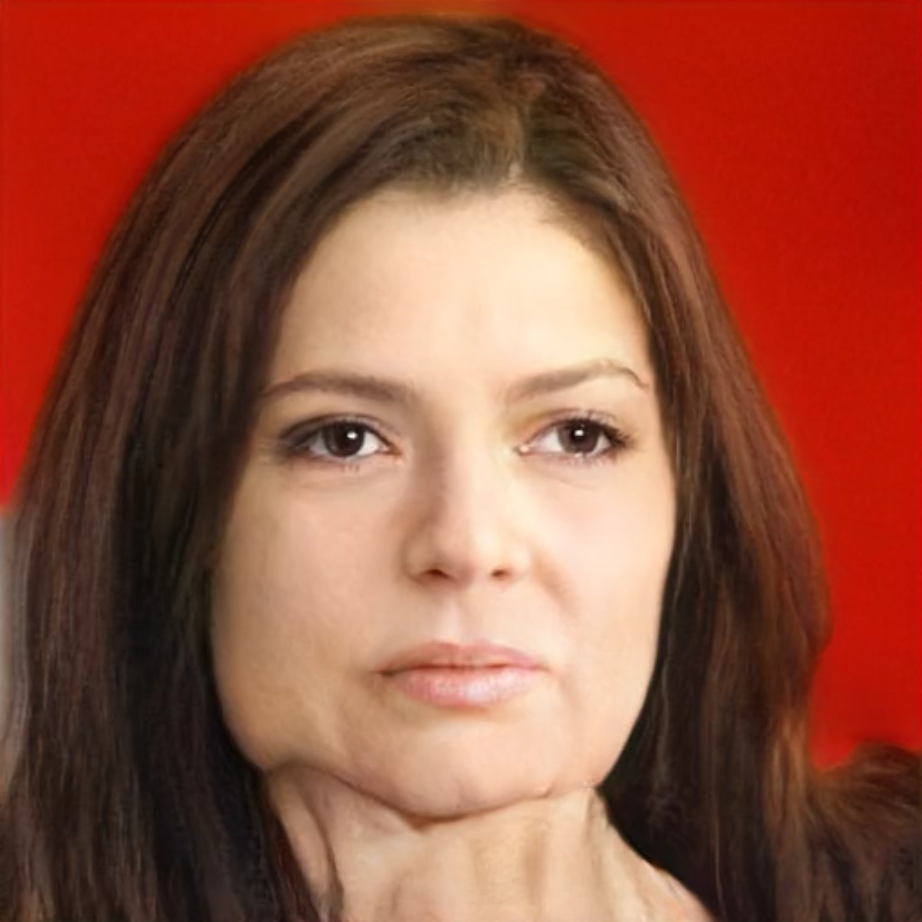} \\
         \begin{tabular}{c}
              Ours \\
              \\
              \\ 
         \end{tabular} & \includegraphics[scale = 0.125]{Experiments/Comparison/985.png} & \includegraphics[scale = 0.5]{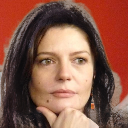} & \includegraphics[scale = 0.5]{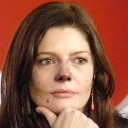} & \includegraphics[scale = 0.5]{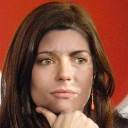} & \includegraphics[scale = 0.5]{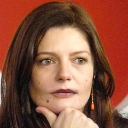} & \includegraphics[scale = 0.5]{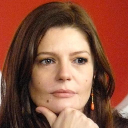} \\
         \begin{tabular}{c}
              SemanticStyleGAN \\
               \\
              \\ 
         \end{tabular} & \includegraphics[scale = 0.125]{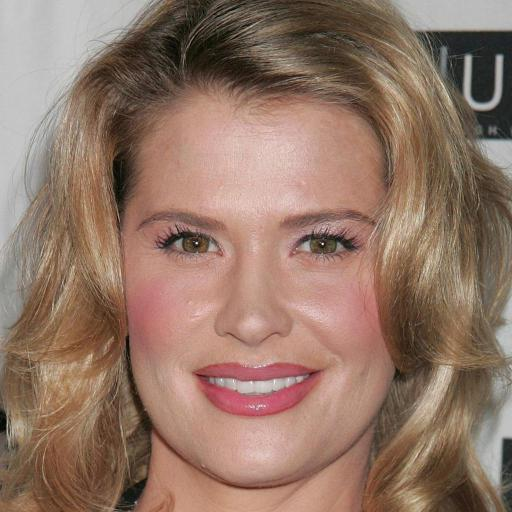} & \includegraphics[scale = 0.07]{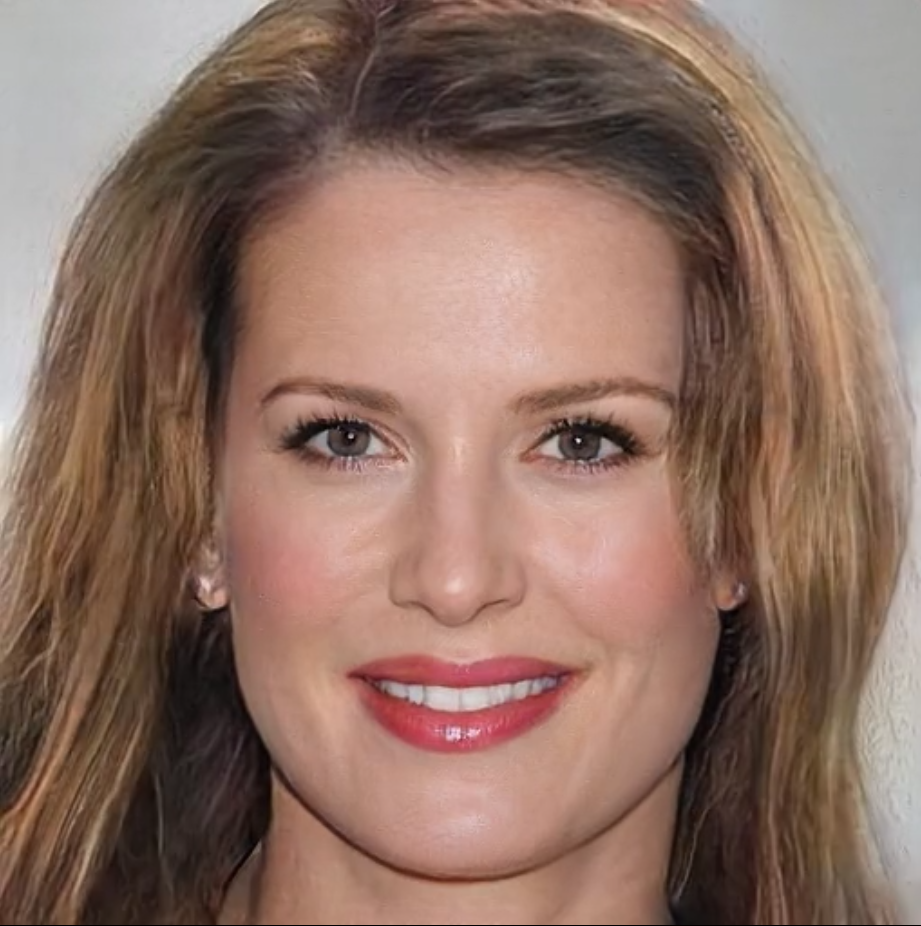} & \includegraphics[scale = 0.07]{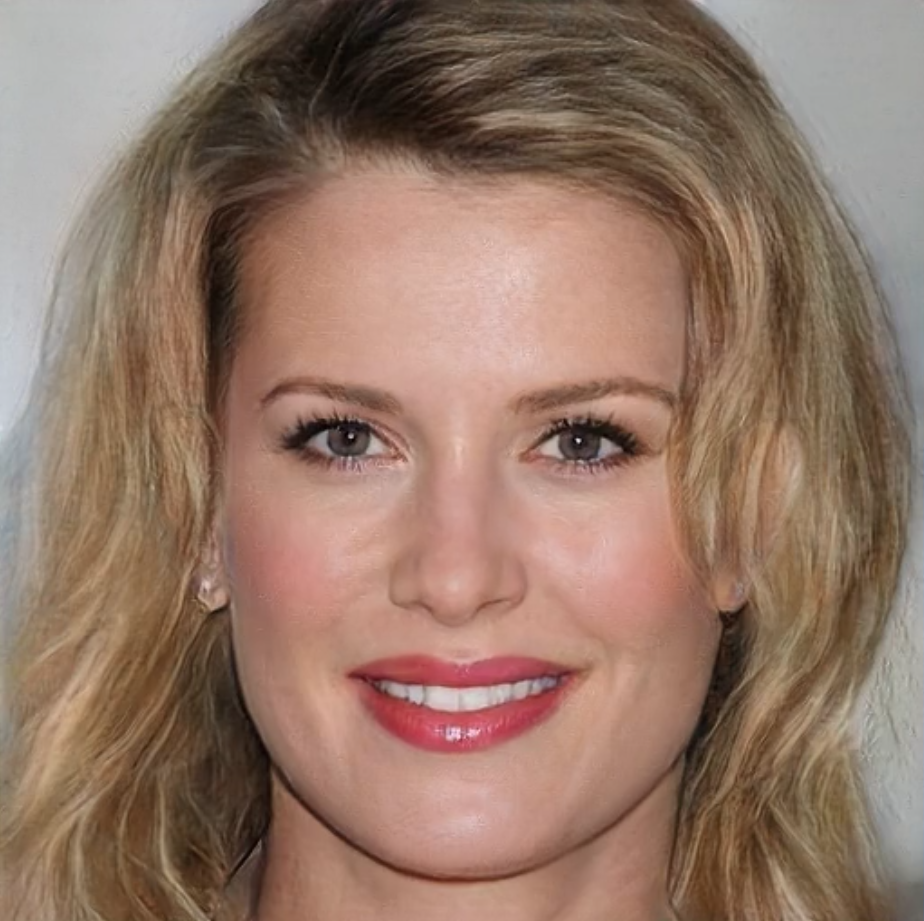} & \includegraphics[scale = 0.07]{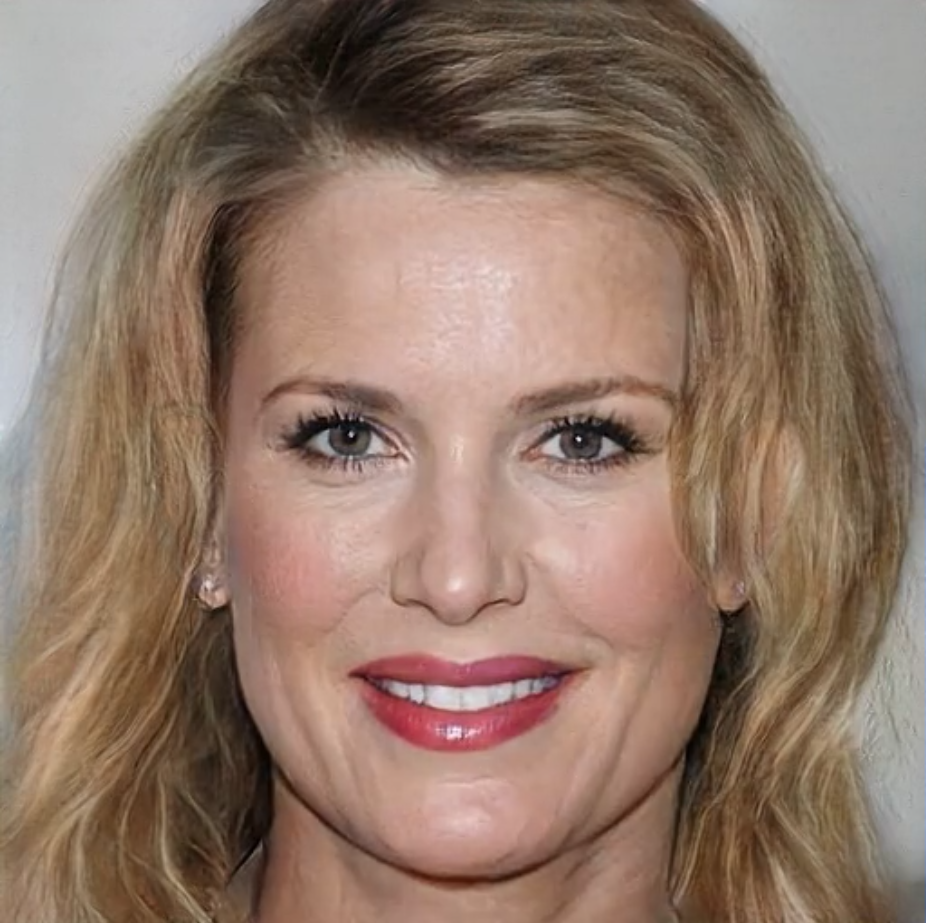} & \includegraphics[scale = 0.07]{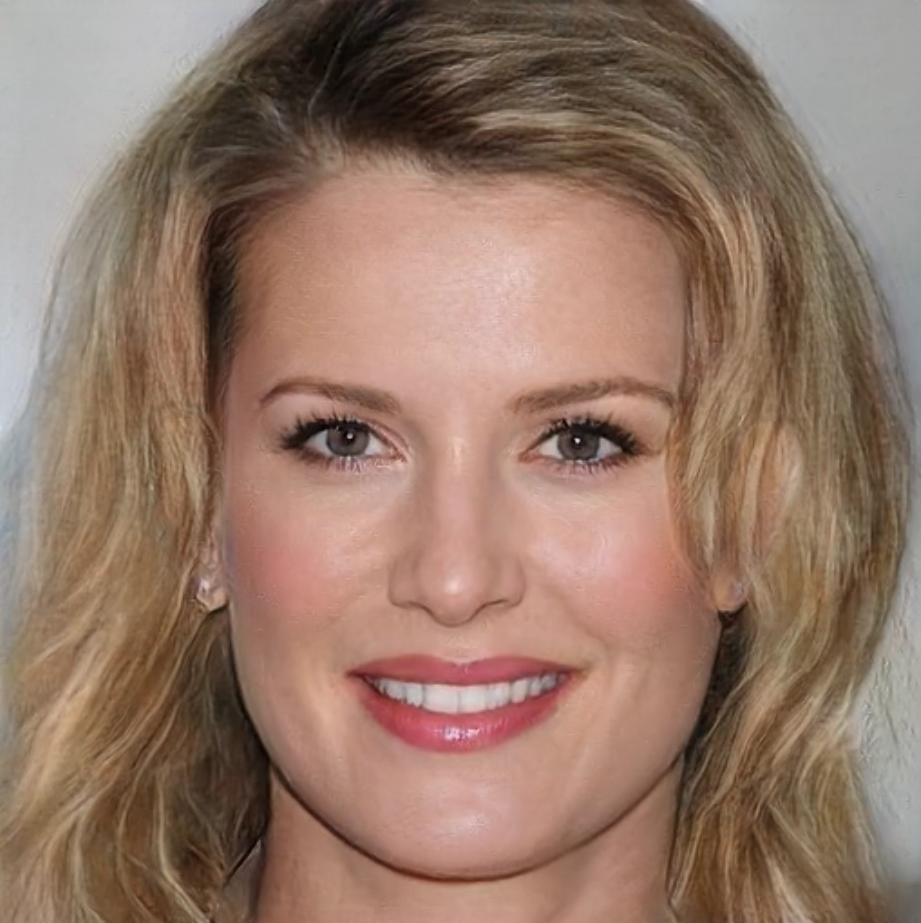} & \includegraphics[scale = 0.07]{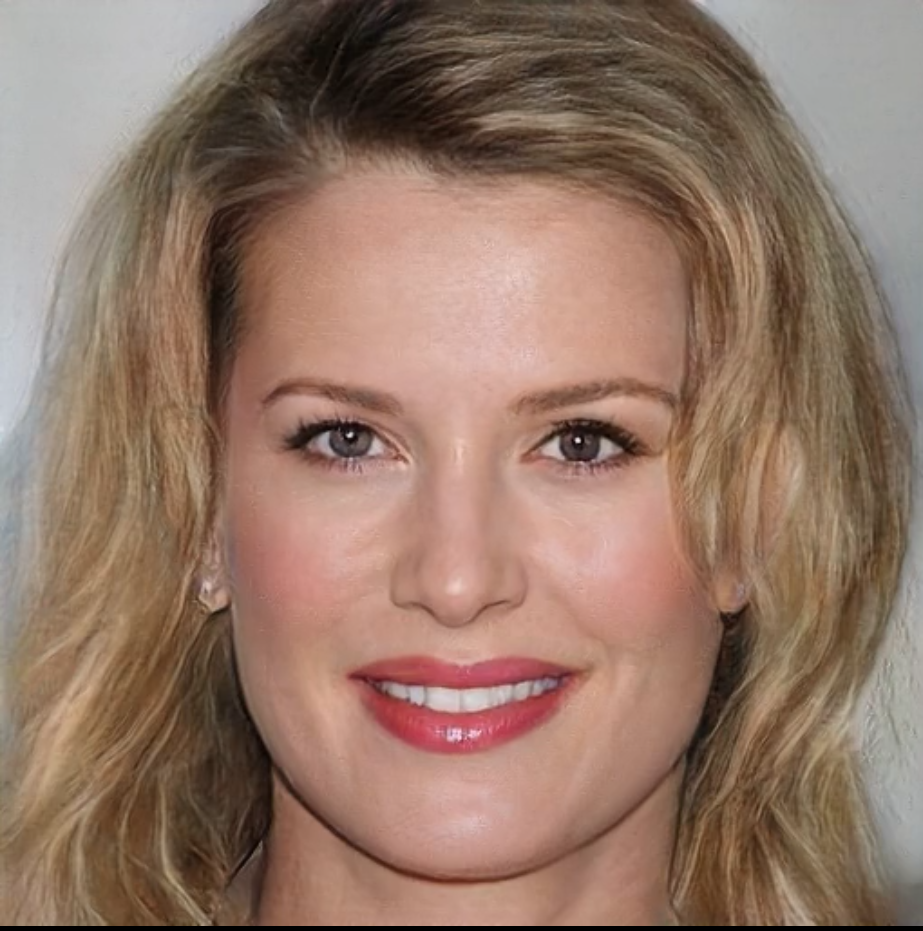} \\
         \begin{tabular}{c}
              Ours \\
               \\
              \\ 
         \end{tabular} & \includegraphics[scale = 0.125]{Experiments/Comparison/2098.png} & \includegraphics[scale = 0.5]{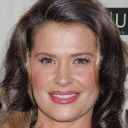} & \includegraphics[scale = 0.5]{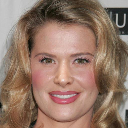} & \includegraphics[scale = 0.5]{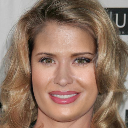} & \includegraphics[scale = 0.5]{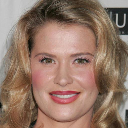} & \includegraphics[scale = 0.5]{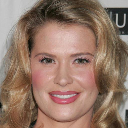} \\
         \begin{tabular}{c}
              SemanticStyleGAN \\
               \\
              \\ 
         \end{tabular} & \includegraphics[scale = 0.125]{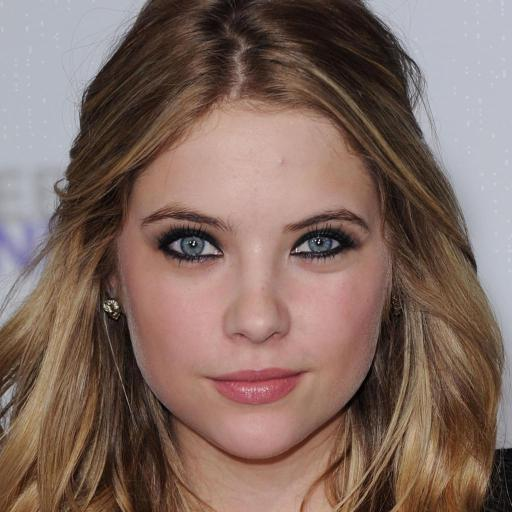} & \includegraphics[scale = 0.07]{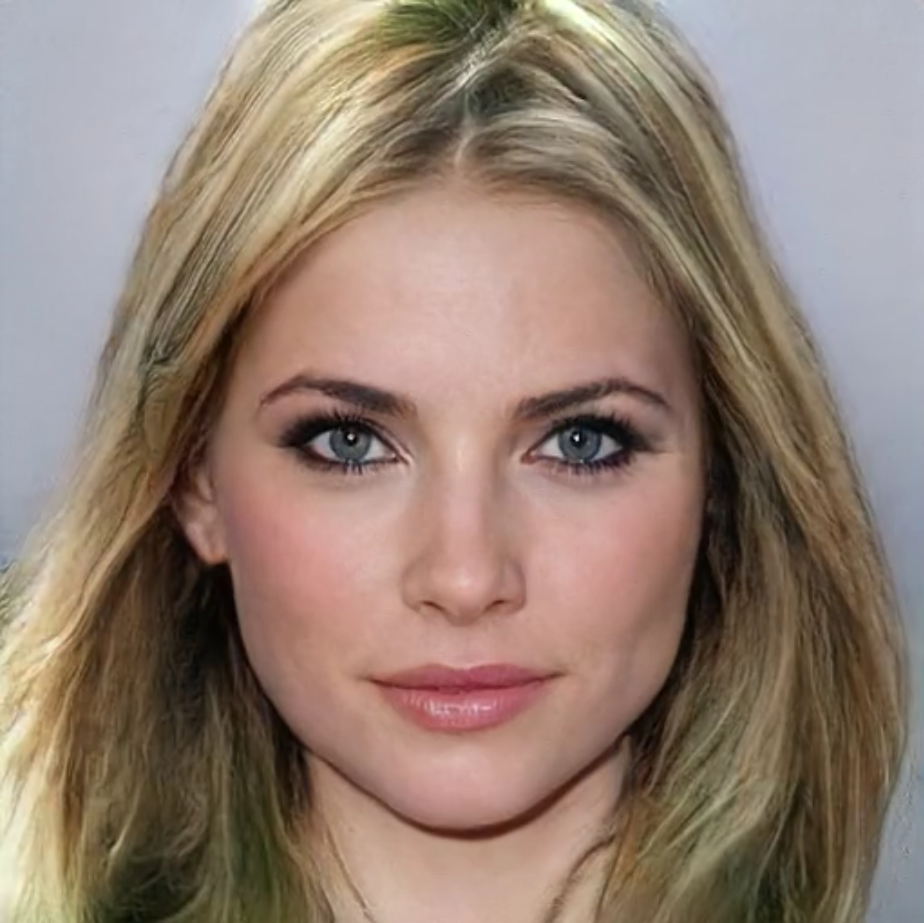} & \includegraphics[scale = 0.07]{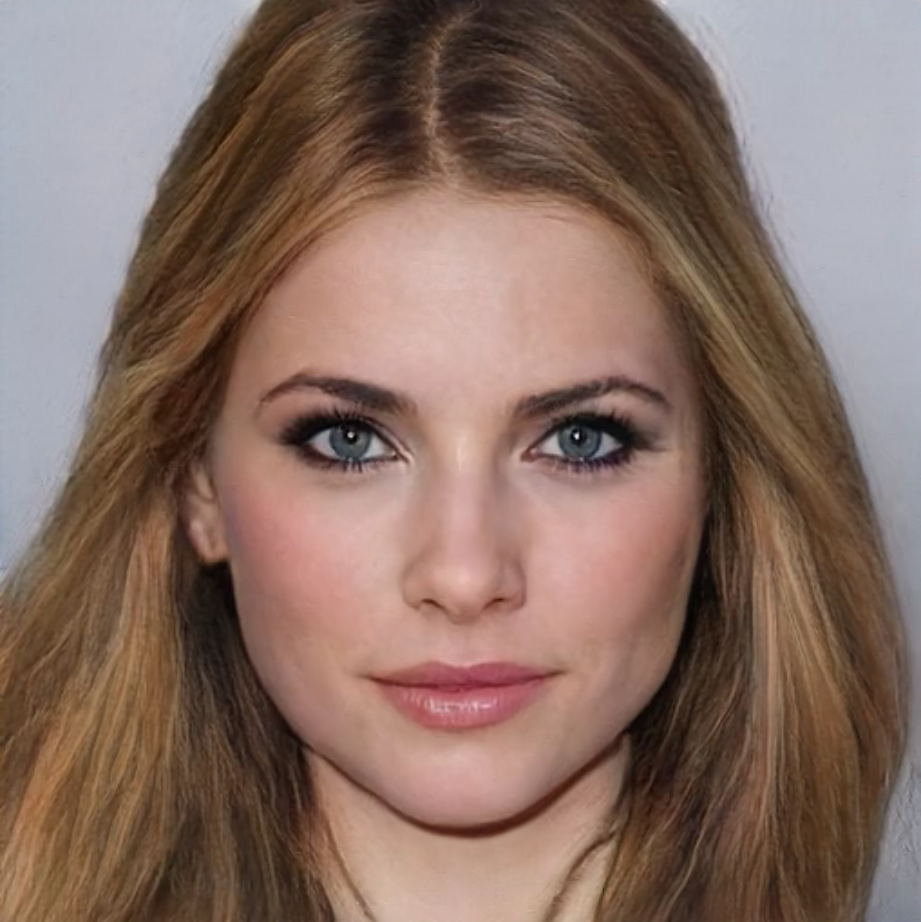} & \includegraphics[scale = 0.07]{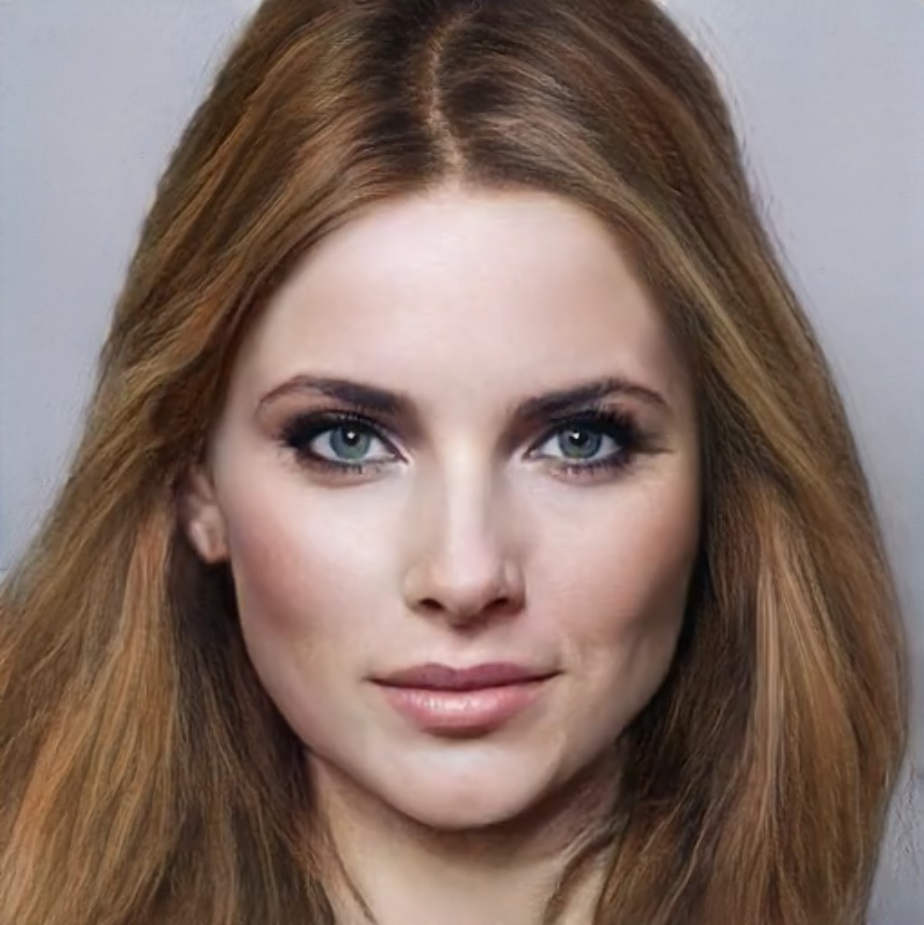} & \includegraphics[scale = 0.07]{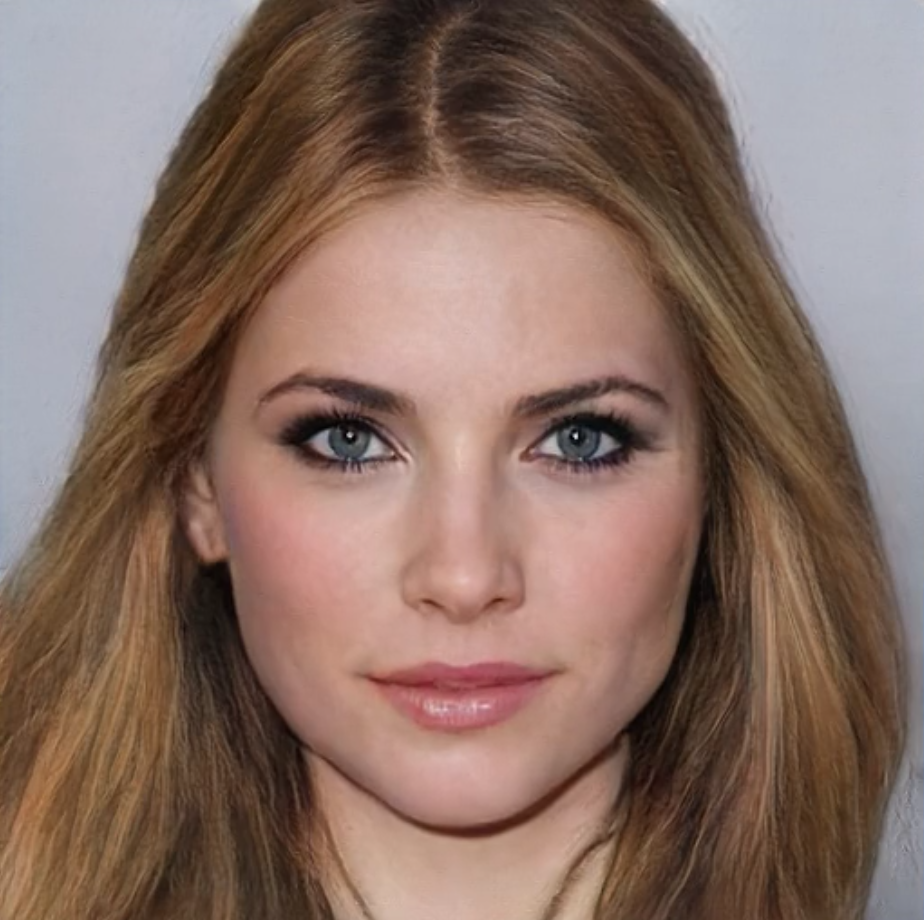} & \includegraphics[scale = 0.07]{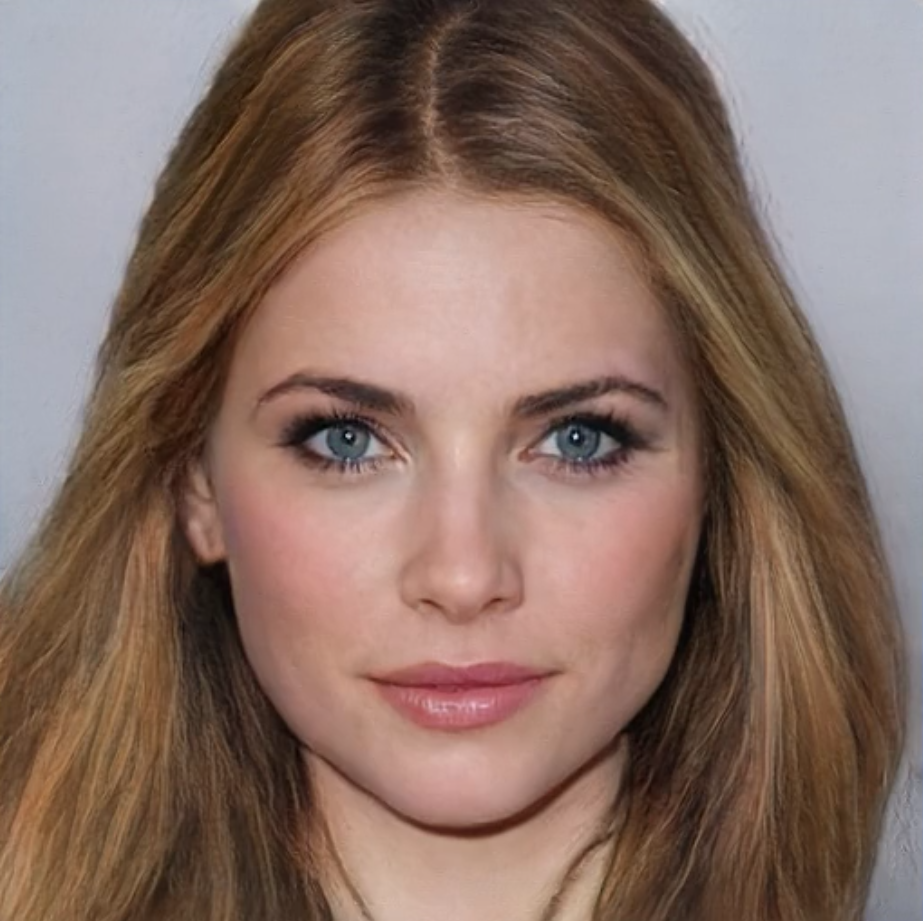} \\
         \begin{tabular}{c}
              Ours \\
              \\
              \\
         \end{tabular} & \includegraphics[scale = 0.125]{Experiments/Comparison/2797.png} & \includegraphics[scale = 0.5]{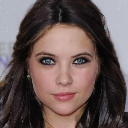} & \includegraphics[scale = 0.5]{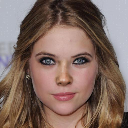} & \includegraphics[scale = 0.5]{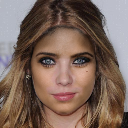} & \includegraphics[scale = 0.5]{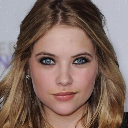} & \includegraphics[scale = 0.5]{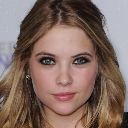} \\
\end{tabular}}
\caption{A comparative analysis of our qualitative results for Localized Style Editing (Texture Manipulation), with respect to SemanticStyleGAN \cite{Shi_2022_CVPR}. Our method is better at preserving the input image structure. Moreover, it performs more localized and pronounced edits.}
\label{fig: qual_compar}
\end{figure}

Fig. \ref{fig: just_qual} highlights the qualitative results in terms of segmentation maps predicted by the SMPN and the ROI-specific style edits obtained in tandem with the SMN. Fig. \ref{fig: qual_compar} (Appendix-A) shows ROI-wise style edits obtained by our mehtod in contrast with those obtained by the SemanticStyleGAN \cite{Shi_2022_CVPR}. It is evident that the SemanticStyleGAN compromises with structure retention from input images in its texture editing pipeline. Since, our method emaerges from the SAE \cite{park2020swapping}, it does not face any issues with structure retention. Moreover, the edits obtained by the SemanticStyleGAN aren't as localized as ours. Editing a region using SemanticStyleGAN affects multiple other regions as well. We produce more noticeable edits as well.
\section{Conclusion}
In conclusion, this work presents a framework for performing structure-preserving, localized, and photorealistic style edits on face images, which agree with the global style scheme of the input image. The presented method does not require any additional human supervision post training and also does away with the need for a computationally expensive iterative-optimization-based latent-inversion process. Performing localized style edits in the presence of occlusions over ROIs is a challenging test scenario for our method. Our method may be used for interesting applications in AR/VR (digital humans) and medicine (dermatology). However, it might find few potentially harmful applications in much the same manner as deepfakes and the likes.
\subsection{Acknowledgement}
Support from Institute of Eminence (IoE) project No. SB22231269EEETWO005001 for Research Centre in Computer Vision is gratefully acknowledged.
\bibliography{anonymous-submission-latex-2023.bib}
\clearpage
\onecolumn
\section{Appendix}
\subsection{A. Qualitative Comparisons}
\end{document}